\documentclass[journal]{IEEEtran}
\usepackage{amsmath,amsfonts}
\usepackage{array}
\usepackage[caption=false,font=footnotesize,labelfont=sf,textfont=sf]{subfig}
\usepackage{textcomp}
\usepackage{stfloats}
\usepackage{url}
\usepackage{verbatim}
\usepackage{graphicx}
\usepackage{tabularx}

\usepackage{xcolor, colortbl}
\usepackage{multirow}
\usepackage[most]{tcolorbox}
\usepackage{booktabs}

\usepackage{cite}
\usepackage{algorithm}
\usepackage{algorithmic}

\usepackage{xspace}
\let\oldtablename\tablename
\renewcommand{\tablename}{\oldtablename\xspace}\usepackage{xspace}
\let\oldfigurename\figurename
\renewcommand{\figurename}{\oldfigurename\xspace}
\def\BibTeX{{\rm B\kern-.05em{\sc i\kern-.025em b}\kern-.08em
    T\kern-.1667em\lower.7ex\hbox{E}\kern-.125emX}}
\usepackage{balance}
\begin{document}
\title{MDA: Availability-Aware Federated Learning\\ Client Selection}
\author{Amin Eslami Abyane, Steve Drew, Hadi Hemmati
% \thanks{Manuscript created October, 2020; This work was developed by the IEEE Publication Technology Department. This work is distributed under the \LaTeX \ Project Public License (LPPL) ( http://www.latex-project.org/ ) version 1.3. A copy of the LPPL, version 1.3, is included in the base \LaTeX \ documentation of all distributions of \LaTeX \ released 2003/12/01 or later. The opinions expressed here are entirely that of the author. No warranty is expressed or implied. User assumes all risk.}
\thanks{Amin Eslami Abyane and Steve Drew are with the Department of Electrical and Software Engineering, University of Calgary, Calgary, AB T2N 1N4, Canada (e-mail: amin.eslamiabyane@ucalgary.ca; steve.drew@ucalgary.ca).}
\thanks{Hadi Hemmati is with the Department of Electrical Engineering and Computer Science, York University, Toronto, ON M3J 1P3, Canada (e-mail: hemmati@yorku.ca).}

}

\markboth{Journal of \LaTeX\ Class Files,~Vol.~18, No.~9, September~2020}%
{How to Use the IEEEtran \LaTeX \ Templates}

\maketitle

%%
%% The abstract is a short summary of the work to be presented in the
%% article.
\begin{abstract}
Recently, a new distributed learning scheme called Federated Learning (FL) has been introduced. FL is designed so that server never collects user-owned data meaning it is great at preserving privacy. FL's process starts with the server sending a model to clients, then the clients train that model using their data and send the updated model back to the server. Afterward, the server aggregates all the updates and modifies the global model. This process is repeated until the model converges. 

This study focuses on an FL setting called cross-device FL, which trains based on a large number of clients. Since many devices may be unavailable in cross-device FL, and communication between the server and all clients is extremely costly, only a fraction of clients gets selected for training at each round. 

In vanilla FL, clients are selected randomly, which results in an acceptable accuracy but is not ideal from the overall training time perspective, since some clients are slow and can cause some training rounds to be slow. If only fast clients get selected the learning would speed up, but it will be biased toward only the fast clients' data, and the accuracy degrades.
Consequently, new client selection techniques have been proposed to improve the training time by considering individual clients' resources and speed. 

This paper introduces the first availability-aware selection strategy called MDA. The results show that our approach makes learning faster than vanilla FL by up to 6.5\%. 
Moreover, we show that resource heterogeneity-aware techniques are effective but can become even better when combined with our approach, making it faster than the state-of-the-art selectors by up to 16\%. Lastly, our approach selects more unique clients for training compared to client selectors that only select fast clients, which reduces our technique's bias. 
The data that support the findings of this study are available in our repository at \url{https://github.com/aminesi/FedML-Extended}.
\end{abstract}

\begin{IEEEkeywords}
Federated Learning, Availability, Reliability, Resource Heterogeneity, Software Engineering
\end{IEEEkeywords}

\section{Introduction}
\label{intro}
With recent advances in mobile devices' hardware and the need for intelligent tasks in mobile applications, federated learning (FL) has emerged as a new learning approach \cite{mcmahan2017communicationefficient} allowing learning from distributed data scattered across multiple clients while preserving their privacy.

A typical FL process consists of multiple rounds of communication between one server and multiple clients. The server broadcasts its most recent global model to the clients in each round. Each client trains the model locally for several epochs and sends its updated trained model parameters to the server. The server collects all updated models received and aggregates them into a new global model for broadcast in the next round. Privacy is preserved as the server only exchanges models with its clients instead of training data, making FL a perfect fit for machine learning scenarios requiring privacy. Recent studies suggest that FL is used in two main settings: cross-device and cross-silo \cite{kairouz2021advances}. In cross-device FL, clients are typically edge devices with limited resources and might not always be available. This study focuses on the reliability and availability concerns of edge devices and discusses the differences between the two settings in the background section.

In cross-device FL, only a fraction of clients gets selected for training in each round to save communication costs and avoid unavailable clients. This introduces a critical topic: client selection strategy \cite{mcmahan2017communicationefficient, kairouz2021advances}. Vanilla FL selects clients randomly between available clients, which has its drawbacks. First, clients have heterogeneous resources which vanilla FL ignores. Second, client availability history is an important factor that should be considered in the selection phase, and vanilla FL does not include that either. In reality, some clients are slower than others, and some are less reliable \cite{hetersurvey, imteaj2021survey}. Considering the resource heterogeneity to make the convergence faster, new client selection techniques have been proposed \cite{fedcs, tifl, oort, powerofchoice, efficienyselection}.
However, none of the proposed techniques consider availability history as a deciding factor. In the FL context, availability is of paramount importance  \cite{hetersurvey}. If a client's availability fluctuates and becomes unavailable in the middle of a training process, the training will likely fail to complete. In a real-world scenario, a timeout is needed to prevent training from getting stuck \cite{bonawitz2019towards}. Therefore, availability, like resource heterogeneity, can impact the training time for FL and must be considered in FL applications.

In this paper, we create three scenarios using different availability settings, namely high, low, and average (for instance, in a low availability scenario, most of the clients are unreliable and unavailable most of the time). In each setting, a mix of clients with different levels of availability are selected from 100k real-world mobile device traces with heterogeneous resource capacities \cite{oort}. We then propose Must Detect the Availability (MDA), an availability-aware client selection strategy aiming to improve FL's performance with availability constraints. We implement these simulations in FedML, an open-source FL framework \cite{fedml}. We run our experiment on two of the most popular datasets used in FL research papers: CIFAR-10 \cite{cifar} and FEMNIST \cite{femnist}.

To show the importance of availability in the training time of FL, we first study the effect of the three availability scenarios in vanilla FL with random client selection. The results show that a low-availability setting can slow down learning time by up to 10\% compared to a high-availability setting, meanwhile degrading the accuracy by up to 5\%.
We then compare our proposed approach to random client selection used in vanilla FL in average- and low-availability scenarios. The results show that MDA can reduce the training time by up to 6.5\%, as well as reduce the number of timed-out rounds by up to 38\%.

To see how our approach works compared to state-of-the-art client selection techniques, we implement two existing client selection techniques proposed in the literature: FedCS \cite{fedcs} and TiFL \cite{tifl}.

The results show that TiFL is more effective than FedCS in improving the training time and gets better results than MDA alone. This is because MDA only considers availability as a selecting factor which has a direct impact on the number of \textit{failed rounds} and an indirect impact on the time. Thus it performs as well or better than baselines in terms of failures but cannot be as fast as them. Since the state-of-the-art techniques select faster clients and directly impact the training time, they are more powerful. However, since availability and resource factors are not overlapping, combining them could produce better results.
Thus we combine TiFL (which shows to be the better option between the baselines) and our approach and call the new technique TiFL-MDA. The results show that TiFL-MDA improves TiFL's speed by up to 16\% and outperforms all the state-of-the-art techniques, and is the fastest in all cases. 

Our contributions are summarized as follows:
\begin{itemize}
    \item We propose an availability-aware client selection for FL called MDA. To the best knowledge, we are the first to consider availability in client selection strategies.
    \item We conduct a complete analysis of client selection techniques in heterogeneous resources and different availability scenarios.
    \item  We propose an availability- and resource-aware technique called TiFL-MDA that outperforms all the baseline client selectors.
\end{itemize}

In the remainder of this paper in Section \ref{background}, we discuss the required background to understand this paper. Section \ref{mda_section} covers our proposed technique. Section \ref{experiments} discusses our goals, experiment design, and results. In Section \ref{related_work}, we cover the most relevant studies, and finally, we conclude our study in Section \ref{conclusion}.

\section{Background}
\label{background}

This section describes the process of FL in ideal conditions and the challenges it has to face in real-world scenarios. Moreover, we discuss the details of the client selection techniques used in this study.

\subsection{Federated Learning}
\label{background_fl}

\begin{figure}
    \centering
    \includegraphics[width=.9\linewidth]{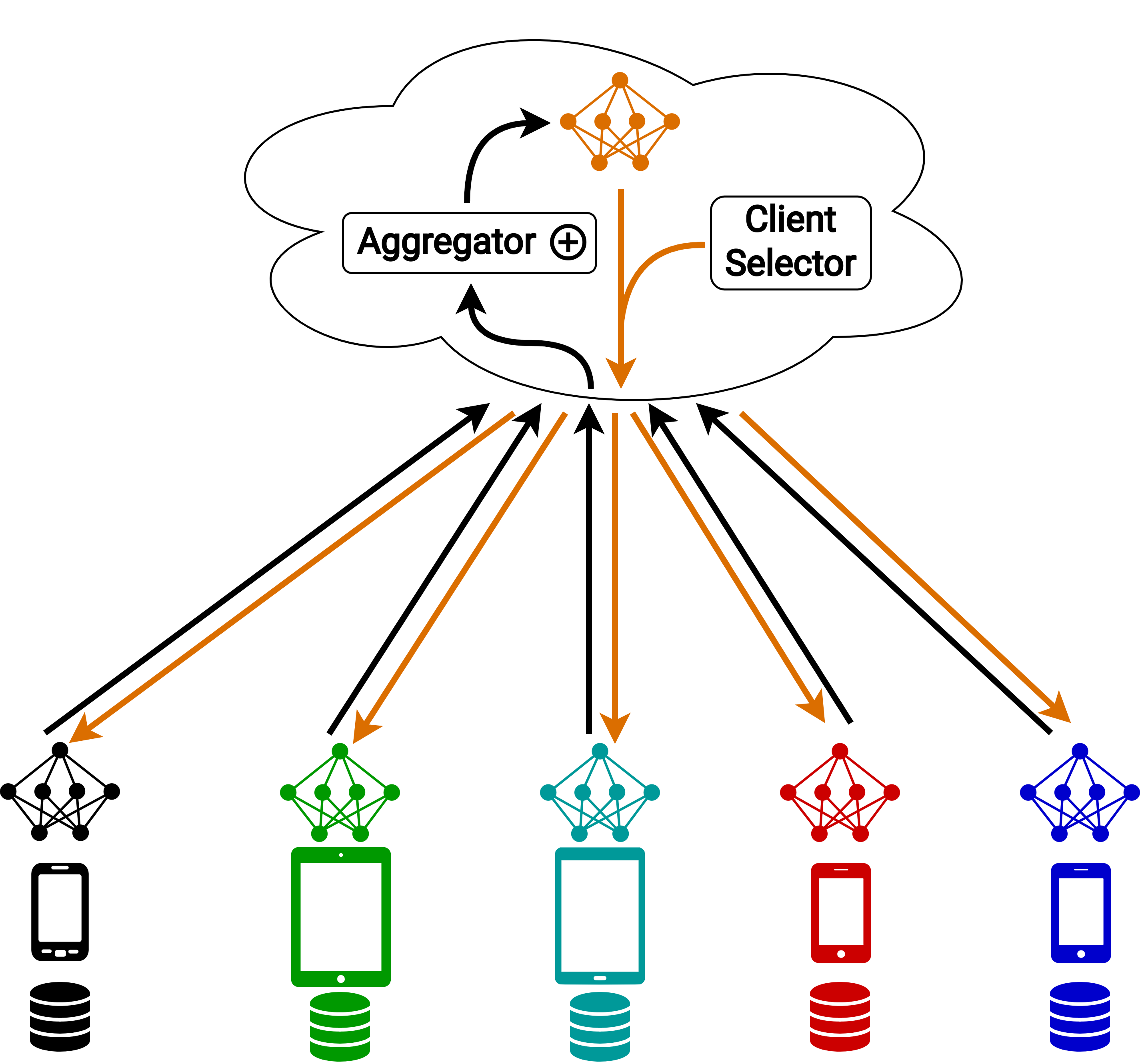}
    \caption{Federated Learning workflow diagram. The orange arrows represent the global model broadcast to the clients, and the black arrows represent client updates sent to the server. Each device has a dataset and a locally trained model, which are represented by different colors.}
    \label{fl_fig}
\end{figure}

Federated Learning (FL) is a new technique introduced in recent years \cite{mcmahan2017communicationefficient}. FL leverages all clients' data to perform distributed learning while ensuring clients' privacy.
\figurename\ref{fl_fig} shows how FL works in practice. FL is an iterative process, and each iteration is called a training round. Each round starts when the server broadcasts a global model to the clients. When a client receives the global model, it performs local training using its data for a few iterations, called local epochs. After a client completes local training, it calculates the difference between the global model received from the server and the trained model. Then it sends the difference called "delta" back to the server. Once the server has all deltas, it will aggregate them to update the global model. This process repeats until the global model converges to a desirable accuracy or the training reaches a certain number of rounds.

This process of FL only transfers model parameters and the delta (difference between global and local models) between clients and the server. Data resides on clients' devices and is never sent to another client or the server, preserving privacy. 

FL was originally designed for mobile devices. Recently, it is applied in other privacy-preserving scenarios. FL can be categorized into two main settings: cross-device and cross-silo \cite{kairouz2021advances}.
In cross-device FL, which is the focus of our study, clients participate in the training process. They are limited in computational and communication resources. They are not reliable and can become unavailable for training. In this setting, only a fraction of clients gets selected for training at each round (The client selector component in \figurename\ref{fl_fig} does this task).
In comparison, cross-silo FL deals with only a few clients, but these clients are highly reliable and powerful. All clients participate in training each round. An example of cross-solo FL is among multiple data centers belonging to different hospitals.

\subsubsection{Real-world challenges}

One of the main challenges in FL (both settings) is the data heterogeneity of clients \cite{mcmahan2017communicationefficient, flsurvey, flsurvey2, kairouz2021advances}. In a real-world case, data distribution is not independent or identically distributed (non-iid), which can cause problems for the FL convergence process. 
Federated Averaging (FedAvg) is shown to be an effective aggregation technique in non-iid cases \cite{mcmahan2017communicationefficient}. FedAvg is the baseline aggregation method in FL, and it is what we use in this study. FedAVg simply takes the mean value of client updates and uses that to update the global model. 

% However, recent studies have introduced other techniques to improve FL performance regarding data heterogeneity \cite{fedprox, fedma}.

% Another important challenge is called byzantine attacks \cite{byzantine, lyu2020privacy, advlens}, aroused when some clients have malicious intentions to disrupt the training process by poisoning their data or model updates. This is a well-studied topic. Various defenses and countermeasures have been proposed to ensure the robustness of FL \cite{krumpaper, medianpaper, foolsgold, emserobustness}. 

The availability of clients and resource heterogeneity \cite{hetersurvey, fedcs, imteaj2021survey} remains a vital challenge in cross-device FL, where clients are less reliable. In this setting, FL uses a client selection technique to select a fraction of clients at each round to make the process withstand the availability problems and reduce communication costs. We will discuss the client selection techniques in the next section.

\subsection{Client selection techniques}
\label{background_selection}

\textbf{Random client selection} \cite{mcmahan2017communicationefficient}:
Client selection is performed randomly in vanilla FL. At each round, the server pings the clients to see which ones are available, creating a pool of available clients. Then the server selects a fraction of them randomly from the pool.
Random client selection is primitive as it only considers availability as a state and does not consider the history of availability, which can be very important. Moreover, resource heterogeneity is not considered. Slow clients may be selected and become the bottleneck of the round.

\textbf{FedCS client selection} \cite{fedcs}: 
This technique is the first and most popular client selection technique that considers resource heterogeneity in its selection process. Like random client selection, it first finds the available clients and selects a fraction of them randomly. Additionally, FedCS has a threshold hyper-parameter and will prevent slow clients from being selected for training. So after the random selection, it performs filtering based on that threshold and only selects a subset of the clients. This parameter determines how fast the overall training will be.

To find clients' required time for training, FedCS gathers information on clients' resources and estimates the time for each client based on model complexity and the client's resources.
The main issue with FedCS is that it excludes many clients for faster training. In non-iid cases, this could have detrimental impacts on the model's accuracy.

\textbf{TiFL client selection} \cite{tifl}:
TiFL (Tier-based Federated Learning) is the last selection technique we study in this paper. Like FedCS, TiFL is a resource-aware technique that aims to speed up the training process.
TiFL starts by putting clients in different tiers based on their speed/resource, gathered before the training process starts. The number of tiers is an important parameter that needs to be selected carefully for the best results.
After tier initialization, in each round, TiFL selects a tier and then selects the desired number of clients only from clients in the selected tier. Tier selection is another important factor in TiFL, and it is performed in a way that faster tiers get selected more than slower tiers.
The reason behind the success of TiFL is two-fold. First, it selects faster tiers more often than slower tiers meaning TiFL is not excluding slow clients, unlike FedCS. Rather, it selects them less often, which makes it fairer. Second, selected clients in a round are from the same tier, so a slow client is never selected with a fast client, therefore avoiding bottlenecks. 

\section{MDA: availability-aware selection}
\label{mda_section}

In this section, we cover the importance of availability in client selection and propose the MDA client selection technique, the first availability-aware client selector to the best of our knowledge. Finally, we propose TiFL-MDA, a combination of TiFL with our MDA technique to have both availability and resource heterogeneity in mind. 

\subsection{Availability in client selection}

Currently, all existing client selectors either do not consider client availability or only consider it as a state during the selection phase \cite{mcmahan2017communicationefficient, hetersurvey, fedcs, tifl, powerofchoice, oort}. 
That means the selector pings all the clients to find available clients at each round and then selects from the available client subset. However, they do not consider the availability history of the clients. For instance, if a client's availability fluctuates and the client gets selected for training, it will likely become unavailable and fails to finish the training process since a real-world FL system must enforce a timeout for each round to withstand potential failures \cite{bonawitz2019towards}.

Timed-out rounds caused by availability fluctuations are never desirable since they slow down the learning process, wasting the resources and energy of the failed clients. An intelligent client selection technique is desired to reduce the occurrence of failed training for more effective model updates.

\subsection{MDA algorithm}

\begin{algorithm}
\caption{MDA algorithm. C: set of all the clients, A: list of availability history for each client, F: list of failure history for each client, r: round index, n: number of clients per round, m: memory length for availability}
\begin{algorithmic}[1] 
\REQUIRE $C, A, F, r, n, m$

\STATE $UpdateHistory(A, F)$

\STATE $Candiadates \leftarrow GetAvailableClients(C, r)$
% \STATE $Failure \leftarrow GetFailedClients(r-1)$

% \FOR{$\textbf{each}\ c \in C$}
% \STATE $A[c].append(c \in Available)$
% \IF{$c \in Failure$}
% \STATE $A[c].append(r-1)$
% \ENDIF
% \ENDFOR

\STATE $W \leftarrow []$
\FOR{$\textbf{each}\ c \in Candiadates$}
\STATE $W[c] \leftarrow 0.5$

\IF{$Length(A[c]) \geq m$}
\STATE $totalTime \leftarrow 0$
\STATE $availableTime \leftarrow 0$
\FOR{$i \leftarrow Length(A[c])-m+1\ \TO\ Length(A[c])$}
\STATE $elapsedTime \leftarrow CalculateElapedTime(i-1, i)$
\STATE $totalTime \leftarrow totalTime+elapsedTime$
\IF{$A[c][i-1]\ \AND\ A[c][i]$}
\STATE $availableTime \leftarrow availableTime+elapsedTime$
\ENDIF
\ENDFOR
\STATE $W[c] \leftarrow availableTime/totalTime$
\ENDIF

\IF{$Length(F[c]) > 0$}
\STATE $pen \leftarrow 0$
\STATE $maxPen \leftarrow 0$
\FOR{$i \gets 0$ \TO $r-1$}
\STATE $p \leftarrow 1/(r-i)$
\STATE $maxPen \leftarrow maxPen+p$
\IF{$i \in F[c]$}
\STATE $pen \leftarrow pen+p$
\ENDIF
\ENDFOR
\STATE $W[c] \leftarrow W[c]*(1-pen/maxPen)$
\ENDIF
\ENDFOR
\STATE $W \leftarrow NormalizeProbabilities(W)$
\STATE $S \leftarrow WeightedRandomSelection(Candiadates, n, W)$
\RETURN $S$
\end{algorithmic}
\label{mda_algorithm}
\end{algorithm}

To reduce the chances of client failures, our approach considers two main factors per client: availability history and failure history. These historical factors are initially empty lists and are progressively filled as training progresses.
The availability history for a client is defined as a list indexed by the number of rounds. The values in the list are Boolean, indicating the client was available at a specific round.
The failure history is also a list, but it only contains the indexes of the rounds that the client had failed in it before, or empty in case the client had not failed in any of the prior rounds.

Algorithm \ref{mda_algorithm} shows how MDA selects clients at each round. The process starts by updating the history of clients (availability and failure) based on the information available to the selector. Following the original FL client selector, \cite{mcmahan2017communicationefficient}, MDA filters clients to gather only available clients in Line 2, then initiates a list of weights $W$ of the size of \textit{Candiatates} and iterates through \textit{Candiatates} in Line 4.  

Lines 6 to 17 show how MDA utilizes the availability history to alter the selection weight for each client. MDA looks at the \textit{m} most recent items in the availability history to calculate the percentage of time that the client was active. Since the exact time of availability change is unavailable to the server and it only knows if a client was active at a certain round, MDA estimates availability time in Lines 12 and 13. MDA considers the client to be available between two consecutive rounds if it was available in both. In other cases, MDA assumes the client was unavailable in that interval. Then MDA calculates the availability percentage in Line 16. If the history is adequate (Line 6), the client receives a weight between 0 and 1 (the availability percentage). Otherwise, it receives 0.5, which is the default value. 

Lines 18 to 29 show the process of applying the failure history in MDA. MDA penalizes the clients that have had failures, and the penalty is larger if the failure is more recent.
There are two critical variables in this part \textit{pen} and \textit{maxPen}. MDA iterates through all rounds from first to last (Line 21). It then calculates the inverse of the distance between that round and the current round and puts it in the variable \textit{p}. The penalty is calculated this way to give more importance to more recent failures. In Line 23, the penalty \textit{p} is always added to \textit{maxPen} since that is supposed to represent the worst case in which the client has been selected every single round till now and failed at all of them. Then MDA checks to see if the client has failed at that round, and only if that is the case, it adds \textit{p} to the \textit{pen}. Finally, MDA calculates the normalized penalty and penalizes the client's weight based on it.  

When the weights are all calculated, in Line 31, MDA normalizes them so that their sum is equal to one, since they represent probabilities. In the end, MDA performs a weighted random selection between all the \textit{Candidates} and returns the selected clients.

\subsection{TiFL-MDA}

MDA works non-invasively and can be combined with any state-of-the-art client selector to make the resulting selector even more powerful. The resulting selection technique will have both availability and resource heterogeneity in mind. As a result, training time will be faster, and FL will use client resources more efficiently. 
Given that at least in theory, TiFL is better than FedCS, we combine MDA with TiFL to propose the first resource and availability-aware technique. As discussed, TiFL's theoretical advantage is because FedCS excludes slow clients completely, but TiFL still includes them but less frequently.

As discussed in Section \ref{background}, TiFL first assigns clients to different tiers, then performs the random selection (just like vanilla FL but on a subset of clients) from a certain tier. We introduce TiFL-MDA, which replaces the random selection component of TiFL with MDA. 

\section{Experiments}
\label{experiments}

\subsection{Objectives and research questions}
The main goal of this study is to show the importance of availability in cross-device FL and how considering availability can improve the speed of the training process in FL.
To achieve this objective, we will propose the following research questions:

\textbf{RQ1: How does vanilla FL's Random client selection perform in different availability scenarios and heterogeneous clients?}

To analyze Random selection, we simulate three different availability settings: low, average, and high. Furthermore, we simulate client resource heterogeneity based on a real-world dataset \cite{oort}. 
Finally, we analyze vanilla FL in different cases that represent real-world scenarios.

\textbf{RQ2: Can MDA improve Random client selection regarding the speed and failures?}

For this RQ and the following RQs, we use the same simulations used in RQ1, except that high availability is no longer included since it is not as interesting as other cases in this paper's context. We compare MDA considering only the availability factor, only the failure factor, and both. Lastly, we compare it with Random selection to answer this question.

\textbf{RQ3: Are resource-aware baselines more effective compared to Random selection?}

To answer this question, we compare FedCS, TiFL, and Random selection techniques to see how they impact the training speed, client failures, and final model accuracy.

\textbf{RQ4: Can availability and resource heterogeneity be used together as selection factors to achieve better results?}

In this question, as discussed in Section \ref{mda_section}, we combine TiFL with our approach to create the TiFL-MDA technique and study its effectiveness.

\subsection{Design}

This section covers this study's design choices like the dataset, simulation techniques, specific parameters, and evaluation metrics.

\subsubsection{Datasets and models}
\label{dataset_section}

\begin{figure}
    \centering
    \subfloat[CIFAR-10\label{cifar_model_fig}]{\includegraphics[width=0.85\linewidth]{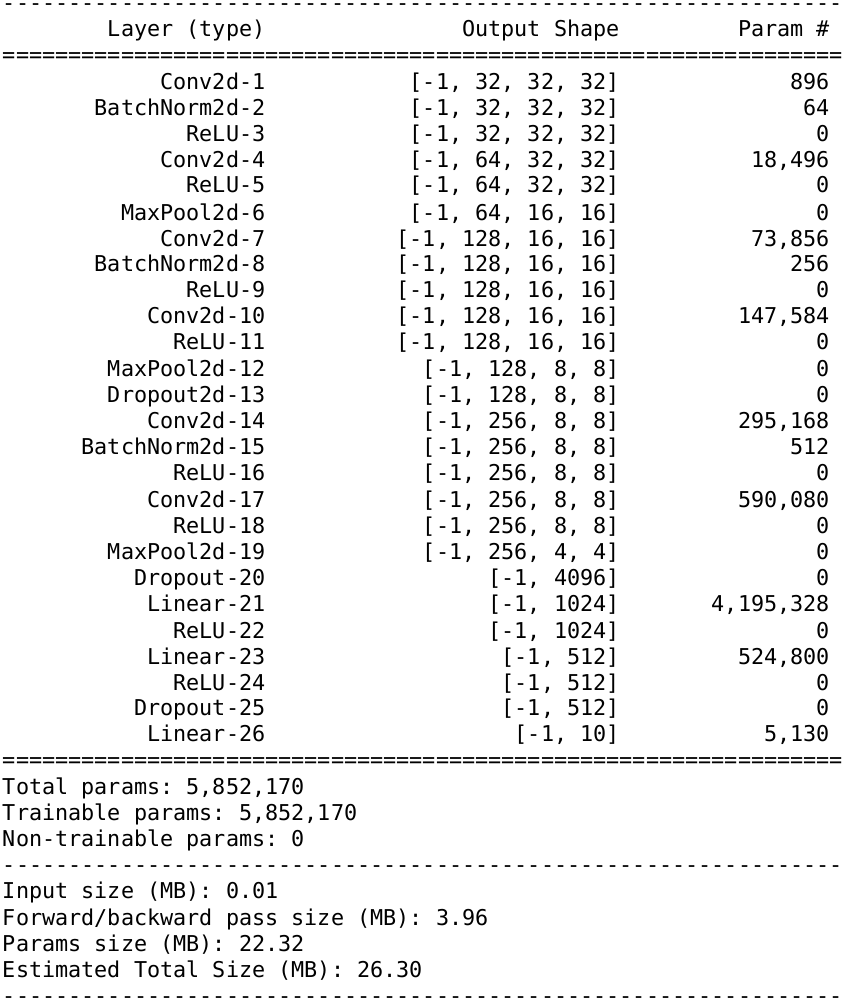}}
    \hfill
    
    \subfloat[FEMNIST\label{femnist_model_fig}]{\includegraphics[width=0.85\linewidth]{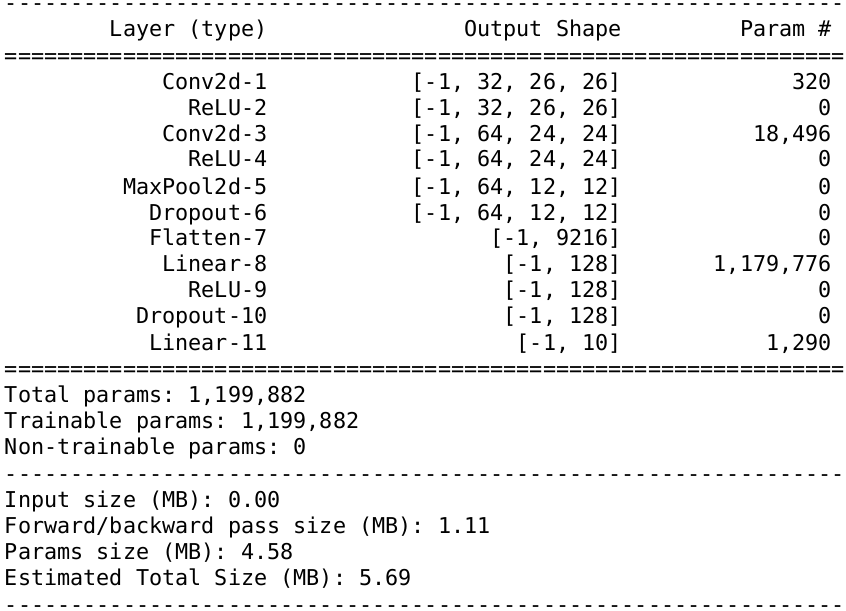}}
    \caption{Summary of models used in this study.}
    \label{model_fig}
\end{figure}

One of the most common tasks in Deep Learning (DL) and FL studies is image classification \cite{adaptivefedopt, mcmahan2017communicationefficient, fedcs, femnist}.
We study two datasets in this paper CIFAR-10 and FEMNIST. These are two of the most studied datasets in the FL community, which is why we use them. 

\textbf{CIFAR-10}:

The CIFAR-10 is an image classification dataset comprising 60,000 32x32 colored images in 10 classes, with 6,000 images per class. The images are split into 50,000 training images, and 10,000 test images \cite{cifar}.
Since the CIFAR-10 dataset is not federated, it must be distributed between clients. To distribute this dataset, we use the Latent Dirichlet Allocation (LDA) technique with an alpha value of 0.2, following the approach used in a related study \cite{adaptivefedopt}. The alpha value can be between zero and one, and the lower it is, the more non-iid the data will be. As a result, this setting ensures a highly non-iid data distribution which would help differentiate selection algorithms in terms of their final accuracy. Another reason for using a non-iid is that in real FL systems, data is distributed in a non-iid manner \cite{kairouz2021advances}.

For the CIFAR-10 dataset, we use a convolutional neural network (CNN) adopted from online sources \cite{cifarmodel}. \figurename\ref{cifar_model_fig} shows a summary of this model. The model consists of six convolutional layers followed by three dense layers. 

\textbf{FEMNIST}:

Federated Extended MNIST (FEMNIST) is a federated version of the Extended MNIST dataset distributed between 3400 clients based on the author \cite{femnist, fedml}. FEMNIST consists of images of lowercase and uppercase letters and digits, resulting in 62 classes. Due to its distribution technique, FEMNIST has a natural non-iid distribution since different authors use different writing styles.

For this dataset, we use a CNN model introduced in a related work \cite{adaptivefedopt}. The model contains two convolutional layers and two dense layers. \figurename \ref{femnist_model_fig} provides a summary of this model. Finally, as seen in the summaries, the CIFAR-10 model is a lot more complex and is nearly four times the size of the FEMNIST model.

\subsubsection{Availability and resource heterogeneity simulation}
\label{sim_section}
Most related studies do not consider availability at all and do not simulate it. However, they consider heterogeneity using techniques like probabilistic distributions with different mean and variance to simulate different speeds \cite{fedcs, tifl}.
Though this type of simulation simulates a resource heterogeneous environment, it does not represent real-world cases like when clients are mobile devices.

To ensure the simulation reflects real mobile devices, new simulation datasets have been proposed that contain availability and resource information about clients gathered from real devices \cite{oort, hetsim}.
Yang et al. propose a dataset that only contains hardware capabilities for three devices \cite{hetsim}, which is extremely limiting and cannot represent a large number of clients, so we do not use this dataset.

However, Lai et al. introduce a comprehensive dataset for simulation purposes that is large-scale and generalizable, and we use it for our simulations \cite{oort}. This dataset includes 107,749 availability traces, each representing a real client. These traces show if a client is available or unavailable at any given time. Furthermore, it contains 500,000 traces that show the hardware capabilities of the clients, like computation speed and communication speed. 

We simply use hardware capacity data from the simulation dataset to simulate resource heterogeneity and assign it to clients.
However, we distribute the availability traces between our clients using different strategies to simulate different availability scenarios. This is possible since the number of availability traces is much greater than the number of clients we use for training.
We start by sorting all the availability traces from worst to best based on factors like availability percentage and availability fluctuation. 
To simulate a low availability case, we pick 60\% of the clients from the start of the sorted traces (the ones with the lowest availability), 20\% from the middle of the sorted traces, and 20\% from the end of the sorted traces (the ones with the highest availability).
Simulation of average and high availability cases follows a similar approach. The difference is that 60\% of clients are selected from the average and high availability parts of the traces, respectively.

\subsubsection{FL system setup}

For the CIFAR-10 dataset, we use 500 clients and select 10 clients per round. The local learning rate is set to 0.01, the number of training rounds is 2500, and each round has an 860s timeout. Apart from the round timeout, all the other parameters are taken from a related study which they came up with using searching in parameter space, and they are shown to be effective \cite{adaptivefedopt}. Regarding the timeout, it is set so that no client misses the round deadline because of slow training, so failures are only caused by availability changes cause that is what we are interested in this paper. 
In the FEMNIST dataset, there are a total of 3400 clients, and at each round, 10 clients get selected for training. The local learning rate is set to 0.1, the number of training rounds is 2000, and each round has a 180s timeout. Like the CIFAR-10 dataset, the timeout is set to only cause failures on availability changes, and other parameters are taken from a related study.
Following a related study, for both datasets, the batch size is set to 20, and each client performs 1 local epoch of training per round to achieve the best results \cite{adaptivefedopt}.

As discussed in Section \ref{background_selection}, resource-aware client selectors have different parameters. For FedCS, we set the threshold to be 160s and 300s for FEMNIST and CIFAR-10 datasets, respectively. These numbers were set to ensure FedCS excludes around 25\% of the clients. Based on our evaluation, if we choose very small thresholds, FedCS's accuracy degrades significantly because most of the clients will be excluded.

For TiFL, following the original paper parameters, we use five tiers. We also make the tier selection strategy so that each tier gets selected 40\% more than the next tier (from fastest to slowest) to make TiFL and FedCS comparable in training time.

\subsubsection{Execution setup and environment}

We build our experimental setup on top of one of the most popular FL frameworks, FedML \cite{fedml}. FedML supports all the functionalities that vanilla FL needs and can run in distributed mode. However, since FedML does not support availability and resource heterogeneity simulations, we integrate the simulations discussed in Section \ref{sim_section} into FedML. Furthermore, we implement all the baseline client selectors discussed in section \ref{background_selection} and our proposed method in FedML to be able to evaluate them. All experiments are done in a distributed manner with one server and ten workers on Digital Research Alliance of Canada cluster nodes. The server process runs with 4 CPU (Intel Silver 4216 Cascade Lake @ 2.1GHz) cores, a GPU (NVIDIA V100 Volta with 32G HBM2 memory and 125 TFLOPS computation power), and 16 GB of memory. The worker processes use the same configuration as the server but do not use a GPU.

\subsubsection{Evaluation metrics}

We use multiple metrics to evaluate the results of this study. The most important metrics are \textit{training time}, the number of \textit{failed rounds}, and the final model \textit{accuracy}. The training time is extremely important since it reflects how well client selectors perform to speed up the training process. The number of failed rounds is another important metric that shows how many rounds at least one client failed and caused a timeout in the server. The model accuracy is also important, showing how much a client selector sacrifices accuracy to achieve faster training.

We also use other metrics like the number of \textit{unique participants}, the \textit{total participants}, and the \textit{average failed clients} per round to better explain the results.

\subsection{Results}

\subsubsection{RQ1 results (Random selection in different availability cases):}
\label{rq1_section}
\begin{table}
    \caption{The effect of availability on Random client selection.}
    \centering
    \subfloat[CIFAR-10\label{rq1_cifar_table}]{
    \begin{tabularx}{.9\linewidth}{Xp{.13\linewidth}p{.13\linewidth}p{.13\linewidth}}
\toprule
 & \multicolumn{3}{c}{Availability}\\
 &         High &      Average &          Low \\
\midrule
Training time(s)         & 1,603,522    & 1,709,592    & 1,767,450    \\
Failed rounds          &       655    &     1,039    &     1,200    \\
Accuracy mean          &        76.17 &        75.55 &        72.66 \\
Accuracy std           &         1.37 &         2.70 &         1.60 \\
Average failed clients &         0.30 &         0.52 &         0.69 \\
Unique participants    &       487    &       492    &       391    \\
Total participants     &    24,238    &    23,694    &    23,277    \\
\bottomrule
\end{tabularx}

}

\subfloat[FEMNIST\label{rq1_femnist_table}]{
    \begin{tabularx}{.9\linewidth}{Xp{.13\linewidth}p{.13\linewidth}p{.13\linewidth}}\toprule
 & \multicolumn{3}{c}{Availability}\\
 &       High &    Average &        Low \\
\midrule
Training time(s)         & 236,739    & 250,424    & 253,606    \\
Failed rounds          &     148    &     222    &     250    \\
Accuracy mean          &      84.11 &      84.55 &      79.11 \\
Accuracy std           &       1.84 &       0.61 &       0.83 \\
Average failed clients &       0.08 &       0.12 &       0.14 \\
Unique participants    &   2,780    &   2,788    &   1,900    \\
Total participants     &  19,846    &  19,765    &  19,730    \\
\bottomrule
\end{tabularx}}
    \label{rq1_table}
\end{table}

\begin{figure}
    \centering
    \subfloat[CIFAR-10\label{rq1_cifar_fig}]{\includegraphics[width=0.8\linewidth]{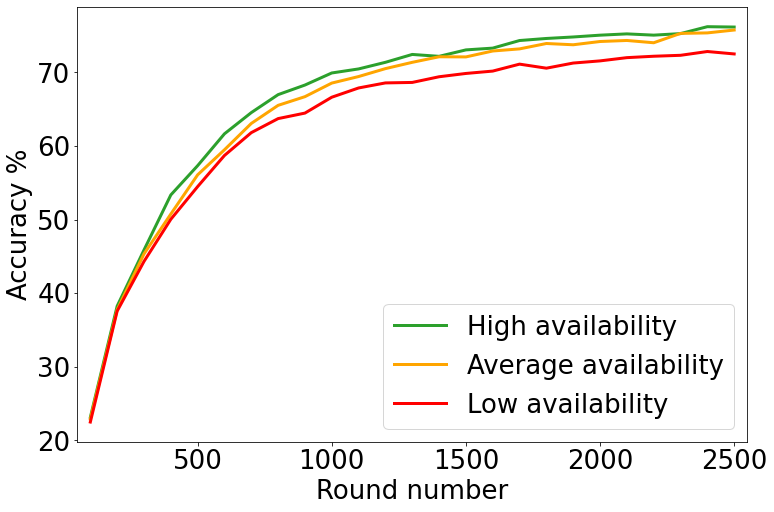}}
    
    \subfloat[FEMNIST\label{rq1_femnist_fig}]{\includegraphics[width=0.8\linewidth]{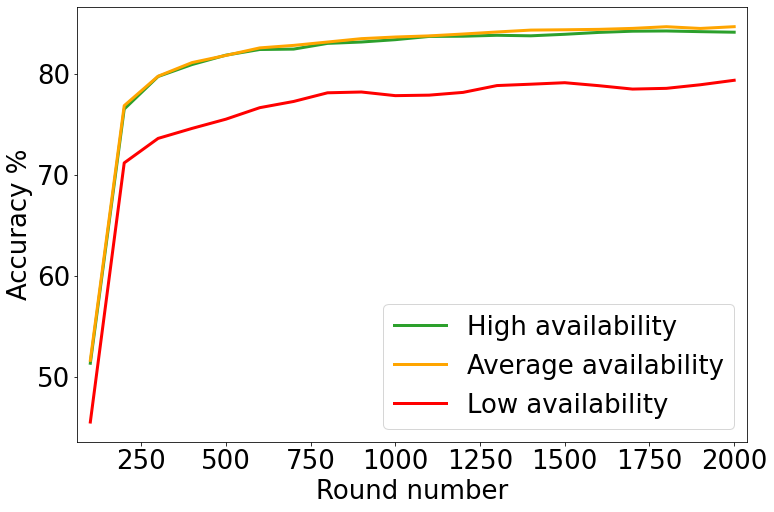}}
    \caption{Convergence plots for Random client selection in different availability scenarios.}
    \label{rq1_fig}
\end{figure}

\tablename \ref{rq1_table} and \figurename \ref{rq1_fig} show the results and convergence plots for this question. As shown in \tablename \ref{rq1_cifar_table} for the CIFAR-10 dataset, the training time increases by around 164,000 seconds (10.2\% change) as availability changes from high to low. This is because the number of \textit{failed rounds} increases by 83.2\%. Note that the time increase is not as significant as the number of \textit{failed rounds} increase since the timeout was set conservatively and close to the slowest client. As the timeout becomes larger, the effect of availability can be seen more significantly on the training speed. 

Furthermore, the \textit{accuracy} also degrades as availability goes lower, which can also be seen in \figurename\ref{rq1_cifar_fig}. the difference between high and average is not significant, but in low availability, we see a noticeable \textit{accuracy} drop, and it is 3.5\% lower than the high availability case. If we look at the \textit{unique participants}, in the high and average cases, it is around 490 clients, which is almost all of the clients (500). But in the low case, it is 391; this, combined with the fact that the dataset is set to be non-iid, is the main reason the accuracy degrades in low availability.

Results for FEMNIST follow a similar pattern, as shown in \tablename \ref{rq1_femnist_table}. 
The FL process is 16,867 seconds slower in the low availability case vs. the high availability case, which is around 7.1\%. Also, in the low availability case, \textit{failed rounds} occur 69\% more than in the high case. Compared to CIFAR-10, these changes are less significant, and that is because the chances of failure are lower in FEMNIST. The first reason is that the fraction of clients selected per round in FEMNIST is less than CIFAR-10 (10/3400 compared to 10/500), while the percentage of low available clients compared to all clients is the same. The other reason is that the task in CIFAR-10 is more complex, as discussed in Section \ref{dataset_section}. Hence, each client's training and communication time is higher, while the availability intervals are the same, so it is more likely in CIFAR-10 that a client goes unavailable before finishing the training (fails).

Moreover, like CIFAR-10 in FEMNIST, the \textit{accuracy} difference between average and high cases is insignificant, and \figurename \ref{rq1_femnist_fig} confirms that. However, in the low availability setting, we see the \textit{accuracy} dropping around 5\% compared to other settings. The reason is the same as discussed in CIFAR-10, and the number of \textit{unique participants} is around 900 less in the low availability case.

Lastly, we see the average client failures per round in low availability cases is 0.69 and 0.14 for CIFAR-10 and FEMNIST, respectively. Considering 10 clients are selected per round, that means 7\% of clients fail to train in the CIFAR-10 dataset, which is 1\% for FEMNIST. This means these clients use energy and resources but do not contribute to the training, so their resources are wasted.  

An important note is that in all cases, in terms of the training time and the number of \textit{failed rounds}, the high availability setting is always a lot better than the other two cases. This means it is closest to the ideal case (no availability failures), and the improvements that an availability-aware selector can produce will be limited, making this setting less interesting from this research's perspective. Thus, we exclude this case from the next RQs and only focus on low and average scenarios.

\begin{tcolorbox}
\textbf{Answer to RQ1:}
The availability is important in FL and can cause serious problems for vanilla FL's Random selections. A low availability setting can slow the FL process compared to a high availability case by 10\% and 7\% for CIFAR-10 and FEMNIST datasets. Furthermore, it can degrade the accuracy by 3.5\% and 5\% for CIFAR-10 and FEMNIST, respectively. 
\end{tcolorbox}

\subsubsection{RQ2 results (MDA client selection):}
\label{rq2_section}

We start answering this question by examining the importance of both factors used in MDA: the availability history and the failure history. To do so, we will evaluate MDA under average availability setting in three configurations: (a) with only failure history, (b) with only availability history, and (c) with both factors.

\begin{table}
    \caption{Results for effect of selection factors on MDA in the average availability case. The fastest configuration is highlighted.}
    \centering
    \subfloat[CIFAR-10\label{rq2_cifar_tabel}]{
    \begin{tabularx}{\linewidth}{Xp{.17\linewidth}p{.23\linewidth}p{.12\linewidth}}
    \toprule
 & \multicolumn{3}{c}{Selector}\\
 &          MDA-failure &  MDA-availability &  MDA \\
\midrule
Training time(s)         &         1,698,820    &              1,619,160    & \cellcolor{gray!25}1,616,575    \\
Failed rounds          &             1,021    &                    679    &       676    \\
Accuracy mean          &                75.74 &                     74.60 &        74.85 \\
Accuracy std           &                 1.66 &                      1.53 &         1.34 \\
Average failed clients &                 0.52 &                      0.32 &         0.31 \\
% Unique participants    &               490    &                    483    &       481    \\
% Total participants     &            23,691    &                 24,206    &    24,220    \\
\bottomrule
\end{tabularx}
    }

    \subfloat[FEMNIST\label{rq2_femnist_tabel}]{
    \begin{tabularx}{\linewidth}{Xp{.17\linewidth}p{.23\linewidth}p{.12\linewidth}}
    \toprule
 & \multicolumn{3}{c}{Selector}\\
 &        MDA-failure &  MDA-availability &  MDA \\
\midrule
Training time(s)         &           254,351   &                251,615    &  \cellcolor{gray!25}248,605    \\
Failed rounds          &               234   &                    202    &      203    \\
Accuracy mean          &                84.44&                     84.47 &       84.37 \\
Accuracy std           &                 0.74&                      0.66 &        0.69 \\
Average failed clients &                 0.12&                      0.11 &        0.11 \\
% Unique participants    &             2,800   &                  2,753    &    2,767    \\
% Total participants     &            19,750   &                 19,789    &   19,788    \\
\bottomrule
\end{tabularx}
    }
    \label{rq2_table}
\end{table}

Results are shown in \tablename \ref{rq2_table}. As it can be seen in \tablename \ref{rq2_cifar_tabel}, for the CIFAR-10 dataset, using only failure history to perform the client selection results in the worst performance. This shows that using the failure history alone is inadequate and does not help the training speed much. However, when availability history is the only deciding factor, the training speed improves compared to failure history by almost 80,000 seconds. The underlying cause for this is the reduction of \textit{failed rounds} from 1,021 to 679, which means there are fewer timeouts, and thus the process is faster. This shows the availability factor is more important than the failure factor, which is expected. Note that we only have failures from availability changes in our experiments, so if there were other failure types, the failure history could also help with those cases.

When combining these factors, results get even better. We see an 8,000s speed-up in the process, and the \textit{average failed client} is 3.1\%, which is lower than other cases. One final observation is that the \textit{accuracy} in failure history is almost 1\% higher than in other cases because it could not effectively filter out problematic clients. Hence, it is more like a random selection, but the speed up we see in the process with MDA using both factors makes this trade-off valuable. 

For the FEMNIST dataset, we see similar patterns in \tablename \ref{rq2_femnist_tabel}. When the failure history is the only deciding factor, we see the worst results with 234 \textit{failed rounds}. The availability history performs much better than the failure history, and the process is almost 2,700 seconds faster with considerably lower \textit{failed rounds}. By combining both factors, MDA performs the best, 3,000 seconds faster than the availability history, which was the best single decoding factor. Although MDA is the fastest, we do not see significant improvements in the \textit{failed rounds} in this case. The speed-up is because the failure history factor implicitly considers clients' speed, so when combined with availability, MDA achieves the best time. 

\begin{table}
    \caption{CIFAR-10 - Results for comparing MDA and Random selector in difference availability cases. The fastest technique is highlighted.}
    \centering
    \begin{tabularx}{\linewidth}{X@{\hskip 5pt}ll|ll}
\toprule
 & \multicolumn{2}{c}{Average}& \multicolumn{2}{c}{Low}\\
 &       Random &          MDA &       Random &          MDA \\
\midrule
Training time(s)         & 1,709,592    & \cellcolor{gray!25}1,616,575    & 1,767,450    & \cellcolor{gray!25}1,651,565    \\
Failed rounds          &     1,039    &       676    &     1,200    &       745    \\
Accuracy mean          &        75.55 &        74.85 &        72.66 &        71.42 \\
Accuracy std           &         2.70 &         1.34 &         1.60 &         1.66 \\
Average failed clients &         0.52 &         0.31 &         0.69 &         0.36 \\
Unique participants    &       492    &       481    &       391    &       350    \\
Total participants     &    23,694    &    24,220    &    23,277    &    24,103    \\  
\bottomrule
\end{tabularx}
    \label{rq2_vsrandom_cifar_table}
\end{table}

\begin{figure}
    \centering
    \subfloat[Average availability\label{rq2_cifar_average_fig}]{\includegraphics[width=0.8\linewidth]{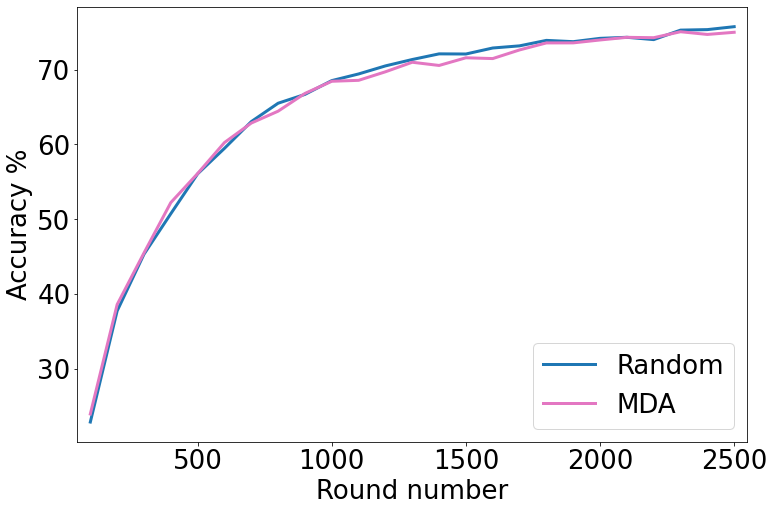}}
    
    \subfloat[Low availability\label{rq2_cifar_low_fig}]{\includegraphics[width=0.8\linewidth]{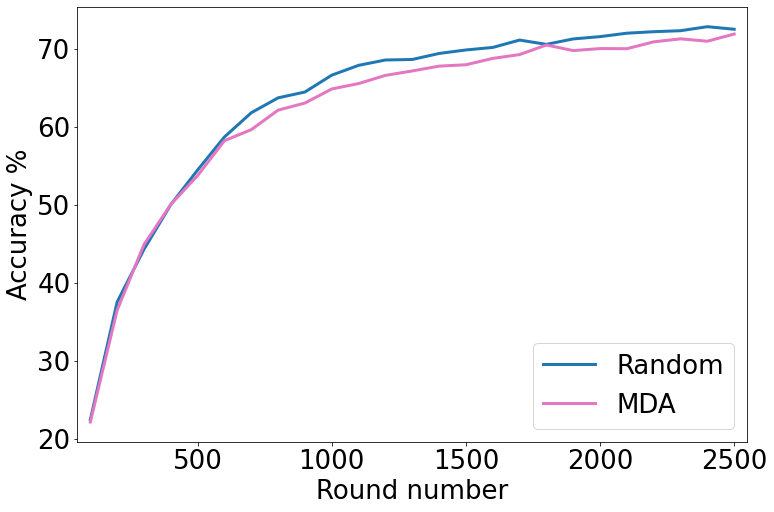}}
    \caption{CIFAR-10 -Convergence plots for MDA and Random selector.}
    \label{rq2_cifar_fig}
\end{figure}

Now that we have established the importance of using both factors in the MDA technique, in the remainder of this question, we compare MDA with Random selection to show the effectiveness of MDA. As discussed at the end of Section \ref{rq1_section}, we only use the average and low availability cases in our experiments since they are more relevant to the study context. 

\tablename \ref{rq2_vsrandom_cifar_table} and \figurename\ref{rq2_cifar_fig} show the results for the CIFAR-10 dataset. As it can be seen in \tablename \ref{rq2_vsrandom_cifar_table}, in the average availability settings, MDA improves the learning process time by 93,000 seconds (5.4\%). Furthermore, it improves the \textit{failed rounds} and reduces it by 363 rounds (34.9\%). This improvement is causing a negligible difference in the final model accuracy, and as the \figurename \ref{rq2_cifar_average_fig} shows, MDA is close to Random in terms of convergence per round. Moreover, the number of \textit{unique 
 participants} is only 10 less than the Random selector, and it has more participants in total since fewer failed clients are in the MDA selector.

Looking at the low availability case, we can see even more improvements in the MDA selector. As the results suggest, MDA improves the total time by 116,000 seconds (6.5\%) which is more significant than the average setting. Also, MDA decreases the \textit{failed rounds} from 1,200 to 745, which is a 38\% improvement. This improvement is, however, more costly in the low availability setting as the \textit{accuracy} drops by almost 1\% and the convergence plot in \figurename\ref{rq2_cifar_low_fig} shows that the Random selector is slightly better there. The main reason behind this is that with low availability, the total number of \textit{unique participants} is lower than the average case (390 compared to 490), and it decreases even more with MDA selection, causing the model not to have all the clients available for training. This trade-off seems reasonable considering the gain in failures and time, but one can always limit the selector's bias to make it fairer.

\begin{figure}
    \centering
    \subfloat[Average availability\label{rq2_femnist_average_fig}]{\includegraphics[width=0.8\linewidth]{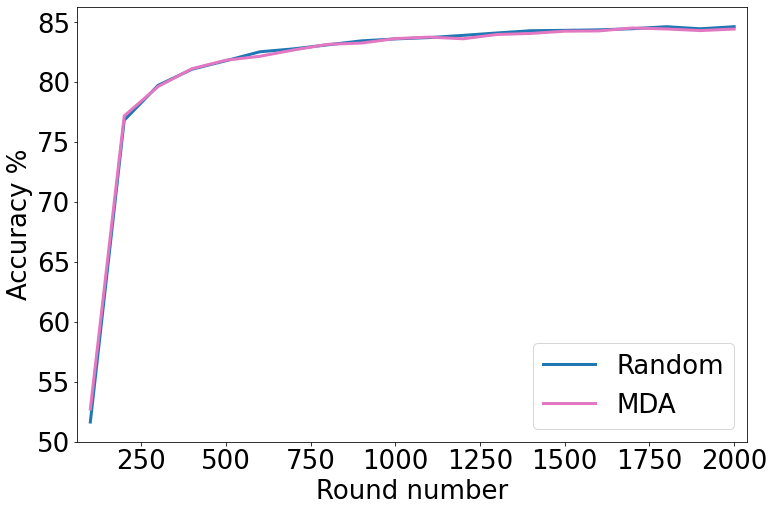}}
    
    \subfloat[Low availability\label{rq2_femnist_low_fig}]{\includegraphics[width=0.8\linewidth]{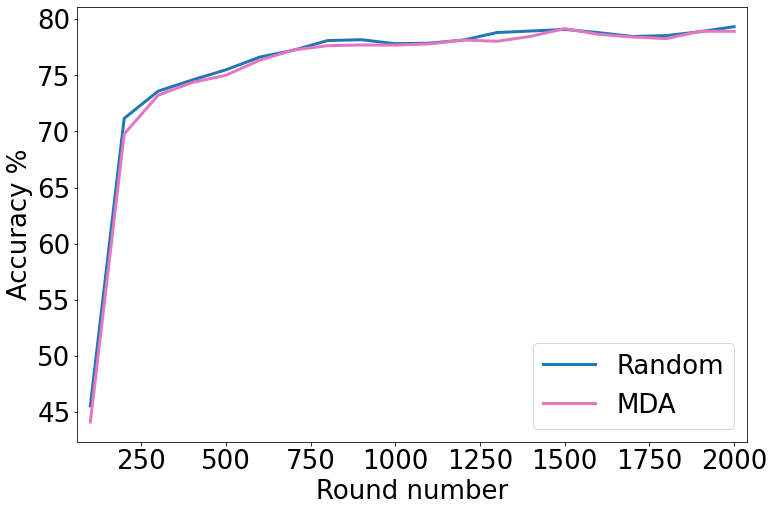}}
    \caption{FEMNIST - Convergence plots for MDA and Random selector.}
    \label{rq2_femnist_fig}
\end{figure}

The results for the FEMNIST dataset are reported in a similar fashion in the \tablename \ref{rq2_vsrandom_femnist_table} and \figurename\ref{rq2_femnist_fig}.
As the results show, in the average case, MDA improves the time by 0.7\%, which is 1.6\% for the low availability case. Furthermore, MDA improves the number of \textit{failed rounds} by 8.5\% and 14.4\% for average and low cases, respectively. These numbers show that MDA shines even more in the low availability case like CIFAR-10, but the improvements are less significant compared to CIFAR-10. 
As discussed before in Section \ref{rq1_section}, in the CIFAR-10 dataset, the chances of failure are higher, and MDA's improvements are more prominent there because there is more chance for improvement compared to FEMNIST.
Lastly, from \figurename\ref{rq2_femnist_fig}, we see that even in the low availability case, MDA and Random are fairly close in accuracy, and improvements in the time do not cause much difference in accuracy. This is because the improvements are not as drastic as in the CIFAR-10 dataset and the data heterogeneity in FEMNIST is natural and less significant compared to the very non-iid distribution we used for the CIFAR-10 dataset.

\begin{table}
    \caption{FEMNIST - Results for comparing MDA and Random selector in difference availability cases. The fastest technique is highlighted.}
    \centering
    \begin{tabularx}{\linewidth}{Xll|ll}
    \toprule
 & \multicolumn{2}{c}{Average}& \multicolumn{2}{c}{Low}\\
 &       Random &          MDA &       Random &          MDA \\
\midrule
    Training time(s)         & 250,424    & \cellcolor{gray!25}248,605    & 253,606    & \cellcolor{gray!25}249,447    \\
    Failed rounds          &     222    &     203    &     250    &     214    \\
    Accuracy mean          &      84.55 &      84.37 &      79.11 &      78.92 \\
    Accuracy std           &       0.61 &       0.69 &       0.83 &       0.89 \\
    Average failed clients &       0.12 &       0.11 &       0.14 &       0.12 \\
    Unique participants    &   2,788    &   2,767    &   1,900    &   1,779    \\
    Total participants     &  19,765    &  19,788    &  19,730    &  19,770    \\
\bottomrule
\end{tabularx}
    \label{rq2_vsrandom_femnist_table}
\end{table}

\begin{tcolorbox}
\textbf{Answer to RQ2:}
Both availability \& failure histories are important to make MDA achieve the best possible results. Also, MDA can improve the total learning process time by up to 6.5\% compared to the Random selector and decrease the number of rounds with a failed client by up to 38\%.
\end{tcolorbox}

\subsubsection{RQ3 results (comparison of baselines):}
\label{rq3_section}

\begin{figure}
    \centering
    \subfloat[Average availability\label{rq3_cifar_average_fig}]{\includegraphics[width=0.8\linewidth]{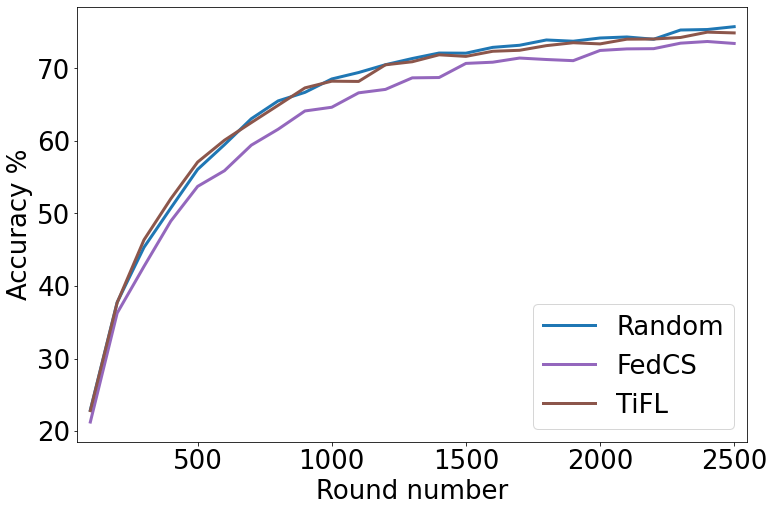}}
    
    \subfloat[Low availability\label{rq3_cifar_low_fig}]{\includegraphics[width=0.8\linewidth]{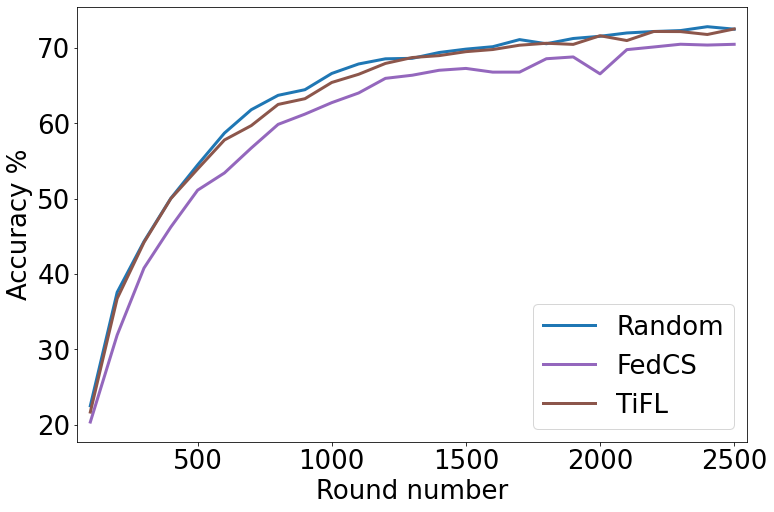}}
    \caption{CIFAR-10 - Convergence plots for baseline selectors.}
    \label{rq3_cifar_fig}
\end{figure}

To answer this question, we report the experiment results in \tablename \ref{rq3_cifar_table} and \figurename \ref{rq3_cifar_fig} for the CIFAR-10 dataset. In the average setting, FedCS improves the Random selector's time by 45.7\% while the TiFL's improvement is 48.7\% and outperforms FedCS slightly. The number of \textit{failed rounds} and \textit{average failed clients} is not an important indicator for these selectors since their goal is not to minimize them. However, they improve these metrics because they use faster clients, which can result in less chance of failure. The interesting metrics that really separate TiFL and FedCS from each other are model \textit{accuracy} and the \textit{unique participants}. As discussed in Section \ref{background_selection}, FedCS ignores slow clients, whereas TiFL selects them less frequently. This can be clearly seen in the \textit{unique participants} as TiFL is almost the same as Random selection with around 490 \textit{unique participants}, but FedCS only trains 360 clients. FedCS's client exclusion manifests perfectly in the model accuracy. As shown in \figurename \ref{rq3_cifar_average_fig}, it clearly is not converging to the best possible results, and TiFL and Random selectors achieve almost 2\% higher accuracy.

The low availability case shows similar results. Regarding the total time, the improvement compared to the Random selector is 42\% and 46.2\% for FedCS and TiFL, respectively. Unlike RQ2, where MDA's results were more significant in the low availability case, both baselines' improvements are better in the average case. This is because these techniques do not consider availability, but MDA is focused on availability, so it controls the client failures better and improves the time better in a low availability setting.
Like the average case, \textit{failed rounds} and \textit{average failed clients} are improved here. An important note is that FedCS is getting fewer client failures than TiFL since it uses only fast clients.
Again, we see from \figurename\ref{rq3_cifar_low_fig} that FedCS is not converging well, and its final model \textit{accuracy} is almost 2\% lower than TiFL.

\begin{table*}
    \caption{CIFAR-10 - Results for comparing baseline selectors. The fastest technique is highlighted.}
    \centering
    \begin{tabularx}{.7\linewidth}{Xp{.06\linewidth}p{.06\linewidth}p{.06\linewidth}|p{.06\linewidth}p{.06\linewidth}p{.06\linewidth}}
    \toprule
 & \multicolumn{3}{c}{Average}& \multicolumn{3}{c}{Low}\\
 &       Random &      FedCS &       TiFL &       Random &      FedCS &       TiFL \\
\midrule
    Training time(s)         & 1,709,592    & 928,120    &  \cellcolor{gray!25}877,288    & 1,767,450    & 1,025,248    &  \cellcolor{gray!25}950,136    \\
    Failed rounds          &     1,039    &     562    &     711    &     1,200    &       714    &     841    \\
    Accuracy mean          &        75.55 &      73.57 &      74.94 &        72.66 &        70.43 &      72.15 \\
    Accuracy std           &         2.70 &       1.92 &       1.17 &         1.60 &         3.23 &       2.70 \\
    Average failed clients &         0.52 &       0.25 &       0.36 &         0.69 &         0.34 &       0.45 \\
    Unique participants    &       492    &     362    &     493    &       391    &       285    &     382    \\
    Total participants     &    23,694    &  17,558    &  24,101    &    23,277    &    17,504    &  23,876    \\
    \bottomrule
\end{tabularx}
    \label{rq3_cifar_table}
\end{table*}

\begin{figure}
    \centering
    \subfloat[Average availability\label{rq3_femnist_average_fig}]{\includegraphics[width=0.8\linewidth]{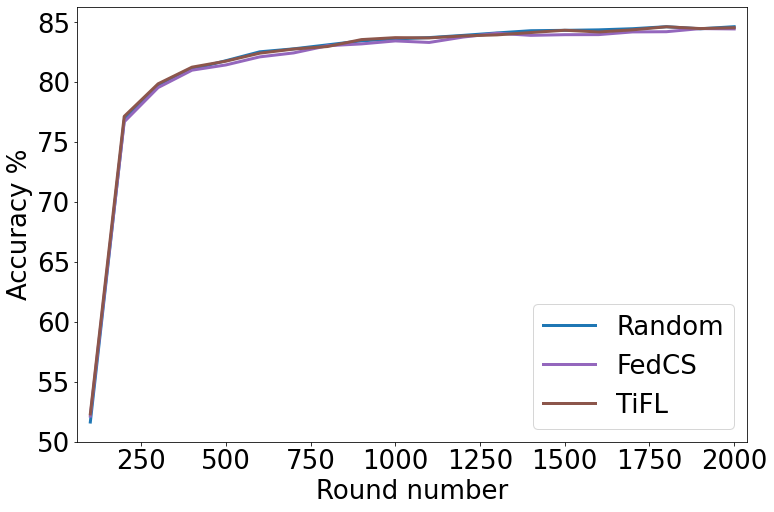}}
    
    \subfloat[Low availability\label{rq3_femnist_low_fig}]{\includegraphics[width=0.8\linewidth]{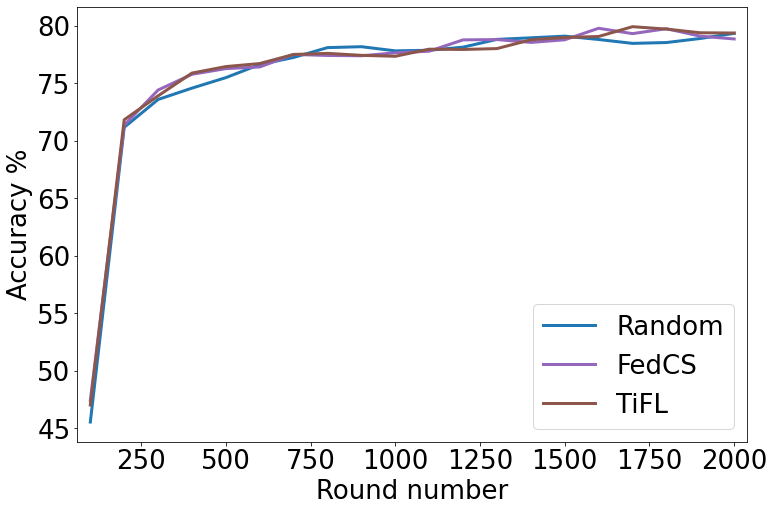}}
    \caption{FEMNIST - Convergence plots for baseline selectors.}
    \label{rq3_femnist_fig}
\end{figure}

\tablename \ref{rq3_femnist_table} and \figurename \ref{rq3_femnist_fig} show the results for the FEMNIST dataset. The average case shows a 54.2\% time improvement for FedCS, and this number for TiFL is 60.2\%, which means in the FEMNIST dataset, TiFL outperforms FedCS as well. Because of the reasons discussed in previous questions, we do not see a significant difference in terms of \textit{accuracy} between any of the selectors, but still, the \textit{unique participants} in FedCS are a lot less than TiFL, meaning TiFL is fairer.
The improvements in the low setting for FedCS and TiFL are 54.1\% and 59.7\%. We also see the same patterns in \textit{failed rounds} and \textit{unique participants}. Furthermore, the \textit{accuracy} differences are insignificant in the low setting, as well.

\begin{table*}
    \centering
    \caption{FEMNIST - Results for comparing baseline selectors. The fastest technique is highlighted.}
    \begin{tabularx}{.7\linewidth}{Xp{.06\linewidth}p{.06\linewidth}p{.06\linewidth}|p{.06\linewidth}p{.06\linewidth}p{.06\linewidth}}
    \toprule
 & \multicolumn{3}{c}{Average}& \multicolumn{3}{c}{Low}\\
 &       Random &      FedCS &       TiFL &       Random &      FedCS &       TiFL \\
\midrule
    Training time(s)         & 250,424    & 114,648    & \cellcolor{gray!25}99,644    & 253,606    & 116,294    & \cellcolor{gray!25}102,386    \\
    Failed rounds          &     222    &     120    &    169    &     250    &     129    &     189    \\
    Accuracy mean          &      84.55 &      84.46 &     84.51 &      79.11 &      78.97 &      79.38 \\
    Accuracy std           &       0.61 &       0.70 &      0.64 &       0.83 &       1.16 &       1.03 \\
    Average failed clients &       0.12 &       0.06 &      0.10 &       0.14 &       0.07 &       0.11 \\
    Unique participants    &   2,788    &   1,985    &  2,553    &   1,900    &   1,389    &   1,797    \\
    Total participants     &  19,765    &  14,834    & 19,809    &  19,730    &  14,602    &  19,789    \\
    \bottomrule
\end{tabularx}
    \label{rq3_femnist_table}
\end{table*}

One final point for this question is that the time improvements we see are far more significant than what RQ2 results showed for the MDA selector. This is expected as TiFL, and FedCS selectors try to select faster clients, which directly impacts the training speed, but MDA tried to improve client failures directly, and the speed up is an indirect effect of that.
Since these are two different approaches (resource-aware selectors and MDA) and do not overlap with each other, combining them may result in the best results, which is the topic of the next question. 

\begin{tcolorbox}
    \textbf{Answer to RQ3:}
    FedCS and TiFL effectively improve the training speed. TiFL can improve the training speed up to 60\%, and FedCS's improvement can be as high as 54\%. In addition, MDA alone has a less significant impact on the speed than TiFL. Lastly, aside from the slight speed advantage that TiFL has over FedCS, it reaches higher accuracies in the end and is the overall superior technique. 
\end{tcolorbox}

\subsubsection{RQ4 results (availability and resource-aware selector):}

To see the effectiveness of TiFL-MDA, we compare it with both TiFL and FedCS techniques. Like previous questions, we first discuss the results of CIFAR-10 and then cover FEMNIST results.

\begin{table*}
    \caption{CIFAR-10 - Results for comparing TiFL-MDA with resource-aware selectors. The fastest technique is highlighted.}
    \centering
    \begin{tabularx}{.7\linewidth}{Xp{.06\linewidth}p{.06\linewidth}p{.073\linewidth}|p{.06\linewidth}p{.06\linewidth}p{.073\linewidth}}
\toprule
 & \multicolumn{3}{c}{Average}& \multicolumn{3}{c}{Low}\\
     &      FedCS &       TiFL &   TiFL-MDA     &      FedCS &       TiFL &   TiFL-MDA\\
\midrule
    Training time(s)         & 928,120    & 877,288    & \cellcolor{gray!25}768,188    & 1,025,248    & 950,136    & \cellcolor{gray!25}799,359    \\
    Failed rounds          &     562    &     711    &     529    &       714    &     841    &     593    \\
    Accuracy mean          &      73.57 &      74.94 &      74.47 &        70.43 &      72.15 &      70.80 \\
    Accuracy std           &       1.92 &       1.17 &       2.46 &         3.23 &       2.70 &       2.29 \\
    Average failed clients &       0.25 &       0.36 &       0.25 &         0.34 &       0.45 &       0.29 \\
    Unique participants    &     362    &     493    &     483    &       285    &     382    &     359    \\
    Total participants     &  17,558    &  24,101    &  24,384    &    17,504    &  23,876    &  24,283    \\
    \bottomrule
\end{tabularx}
    \label{rq4_cifar_table}
\end{table*}

\begin{figure}
    \centering
    \subfloat[Average availability\label{rq4_cifar_average_fig}]{\includegraphics[width=0.8\linewidth]{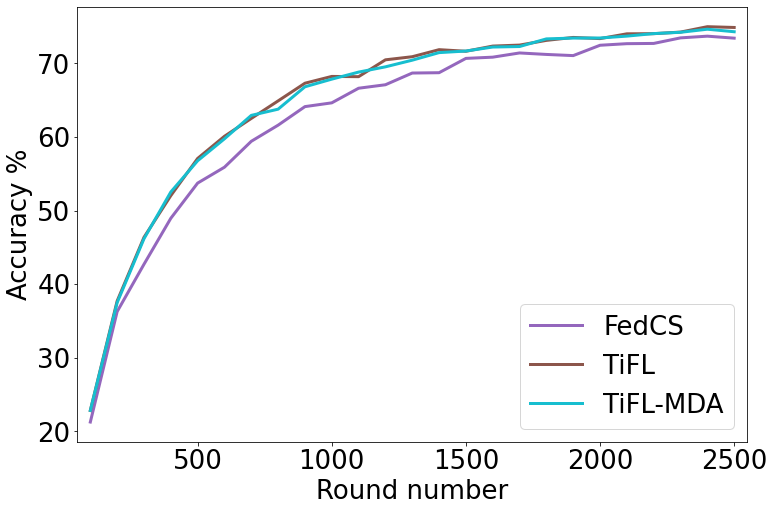}}
    
    \subfloat[Low availability\label{rq4_cifar_low_fig}]{\includegraphics[width=0.8\linewidth]{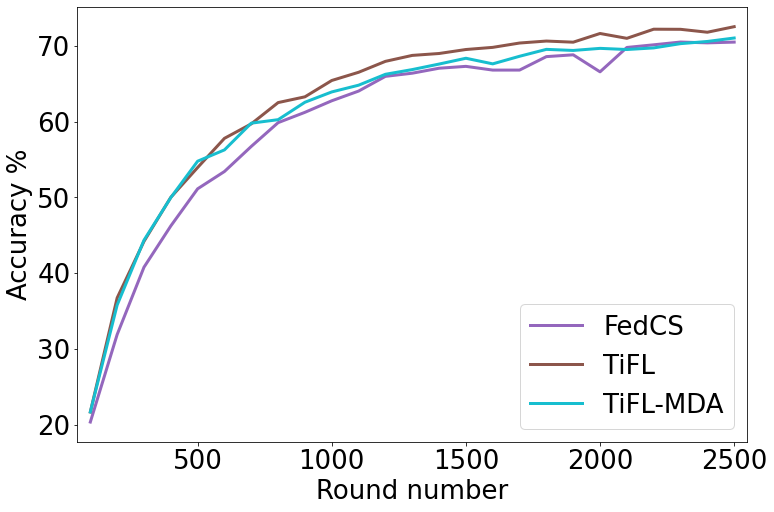}}
    \caption{CIFAR-10 - Convergence plots for TiFL-MDA and state-of-the-art selectors.}
    \label{rq4_cifar_fig}
\end{figure}

\tablename \ref{rq4_cifar_table} and \figurename \ref{rq4_cifar_fig} show the results for the CIFAR-10 dataset. 
In the average availability setting, TiFL-MDA improves TiFL, which was already superior to FedCS. In terms of the learning process speed, TiFL-MDA is 17.3\% faster than FedCS, and it is 12.4\% faster than TiFL. Furthermore, it improves the number of \textit{failed rounds} of TiFL by 182 rounds or 25.5\%. The only advantage FedCS had over TiFL was the \textit{failed rounds}, and TiFL-MDA is even better than FedCS in that regard. Moreover, we see that the model \textit{accuracy} is virtually intact compared to TiFL, and as \figurename\ref{rq4_cifar_average_fig} shows, TiFL-MDA's accuracy is on par with TiFL and is much higher than FedCS.  

Low availability results show similar patterns, and TiFL-MDA can improve state-of-the-art techniques. Compared to FedCS, TiFL-MDA is 22\% faster, and it improves TiFL's training time by 15.8\%. Like RQ2, where MDA had more significant improvements in the low availability case, TiFL-MDA's speedup is more significant in the low availability scenario compared to the average case. Moreover, the number of \textit{failed rounds} of TiFL is improved by almost 250 rounds (30\%), making the TiFL-MDA better than all the studied techniques. Also, TiFL-MDA achieves 3\% \textit{average failed clients} per round, which is the lowest among all the studied techniques.
The only downside that TiFL-MDA has in the low availability case is the \textit{accuracy} drop it experiences compared to TiFL, which is around 1\%. The same issue was seen in RQ2, where MDA was being compared to the Random selector. However, this small drop is justified given its improvement over TiFL, and TiFL-MDA is still better than FedCS in terms of accuracy.

\begin{table*}
    \caption{FEMNIST - Results for comparing TiFL-MDA with resource-aware selectors. The fastest technique is highlighted.}
    \centering
    \begin{tabularx}{.7\linewidth}{Xp{.06\linewidth}p{.06\linewidth}p{.073\linewidth}|p{.06\linewidth}p{.06\linewidth}p{.073\linewidth}}\toprule
 & \multicolumn{3}{c}{Average}& \multicolumn{3}{c}{Low}\\
     &      FedCS &       TiFL &   TiFL-MDA     &      FedCS &       TiFL &   TiFL-MDA\\
\midrule
    Training time(s)         & 114,648    & 99,644    & \cellcolor{gray!25}96,005    & 116,294    & 102,386    & \cellcolor{gray!25}98,621    \\
    Failed rounds          &     120    &    169    &    151    &     129    &     189    &    150    \\
    Accuracy mean          &      84.46 &     84.51 &     84.67 &      78.97 &      79.38 &     79.84 \\
    Accuracy std           &       0.70 &      0.64 &      0.61 &       1.16 &       1.03 &      1.24 \\
    Average failed clients &       0.06 &      0.10 &      0.08 &       0.07 &       0.11 &      0.08 \\
    Unique participants    &   1,985    &  2,553    &  2,532    &   1,389    &   1,797    &  1,758    \\
    Total participants     &  14,834    & 19,809    & 19,832    &  14,602    &  19,789    & 19,841    \\
    \bottomrule
\end{tabularx}
    \label{rq4_femnist_table}
\end{table*}

\begin{figure}
    \centering
    \subfloat[Average availability\label{rq4_femnist_average_fig}]{\includegraphics[width=0.8\linewidth]{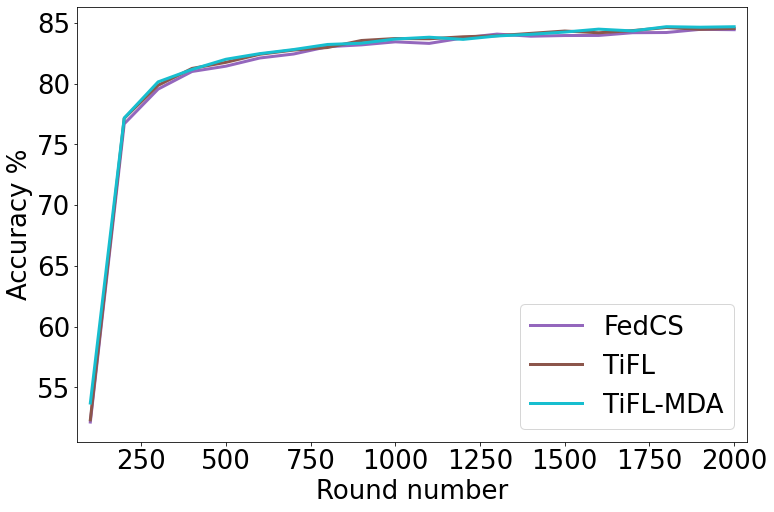}}
    
    \subfloat[Low availability\label{rq4_femnist_low_fig}]{\includegraphics[width=0.8\linewidth]{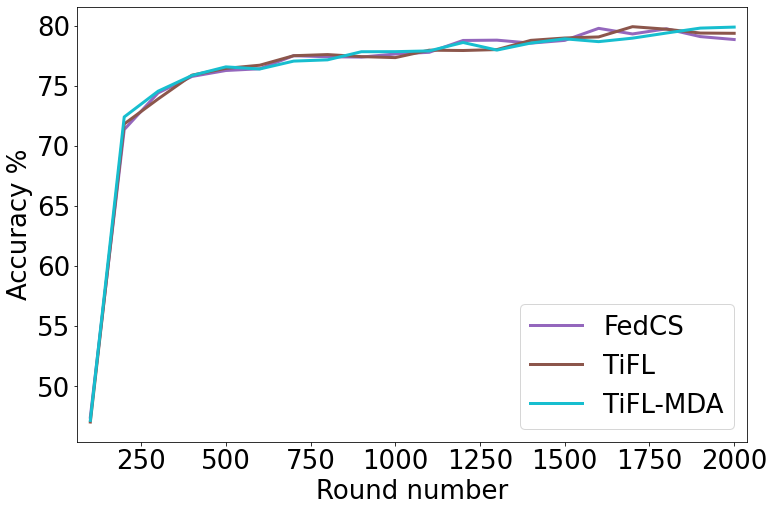}}
    \caption{FEMNIST - Convergence plots for TiFL-MDA and state-of-the-art selectors.}
    \label{rq4_femnist_fig}
\end{figure}

\tablename \ref{rq4_femnist_table} and \figurename\ref{rq4_femnist_fig} report the results of the FEMNIST dataset. As we see in the average case, TiFL-MDA improves the speed of TiFL by 3.6\%, and it makes FedCS faster by 16.2\%. Also, it improves the number of \textit{failed rounds} of TiFL by 20 rounds (11.7\%). As we saw in RQ2 for MDA, TiFL-MDA's improvements in the FEMNIST dataset are less significant than CIFAR-10. So, it cannot quite achieve the \textit{failed rounds} that FedCS has, but it closes the gap. As discussed in previous questions, the \textit{accuracy} differences are not noticeable in the FEMNIST dataset.

When we look at the low availability setting results, TiFL-MDA improves the training time of TiFL and FedCS by 3.7\% and 15.2\%. Furthermore, it improves the \textit{failed rounds} of TiFL by 40 rounds (21\%), getting it closer to FedCS in this metric. Lastly, as the \figurename\ref{rq4_femnist_low_fig} shows, the accuracy differences are insignificant, and all techniques converge similarly.

\begin{tcolorbox}
    \textbf{Answer to RQ4:}
    Having both availability and resource heterogeneity as selection factors can indeed cause better results, as shown for TiFL-MDA. TiFL-MDA can improve the speed of the fastest state-of-the-art technique used in this study (TiFL) by up to 16\%. Furthermore, this improvement caused very small to zero accuracy degradation, making TiFL-MDA the best all-around selector technique for this context.
\end{tcolorbox}

\subsection{Limitations and threats to validity}
%TODO what category does threats to validity of my approach goes to?
One of the limitations of our proposed approach is that it gets slightly biased in the low availability setting. As the results showed, MDA and TiFL-MDA will force the selection of more available clients to reduce the chances of failure, which impacts the accuracy in some cases. This is a trade-off between client failure (and consequently the training time) and the accuracy that may be more optimized when considering fairness factors. We plan to do this in future work.

\textbf{Internal validity:}
To mitigate the internal threats, we used well-studied and popular client selection techniques in our study. We implemented them based on the algorithms provided in their papers, and two of the authors went through the implementations.
Furthermore, we used the FedML framework, a well-known FL framework, to perform our experiments. FedML provides most of the functionalities required for this study, like distributed training, image classification datasets, and the distribution technique used for centralized datasets. We integrated the client selection and resource/availability simulation components into FedML to run our experiments.

\textbf{External validity}
Questions may arise regarding the image datasets, models, and the simulation dataset used in this study.
To mitigate these issues, we used two of the most popular datasets used in FL studies CIFAR-10 and FEMNIST. Our choices of application datasets contain both naturally federated datasets and centralized datasets distributed manually to control the data distribution. 
Regarding the models, we used CNNs used in other studies and public sources, which are shown to be effective. It is possible to use more advanced models to achieve better accuracies, but since the main goal of this study is to analyze training speed and accuracy is not the main goal we used CNNs.
Regarding the simulation of availability and resource heterogeneity, we used a real-world dataset with information from thousands of clients, and we did not use probabilistic simulations, which makes our study more realistic.

\textbf{Construct validity:}
We used total training time as the main metric of this study, following what related studies had done. We also used accuracy as another metric to show the model's performance in the classification task. Furthermore, we used other metrics like \textit{average failed clients}, the number of \textit{failed rounds}, and \textit{unique participants} to justify results and better show the effects of clients' availability.

\section{Related work}
\label{related_work}
In this section, we look at the related surveys and state-of-the-art client selection techniques for FL. 
Lim et al. perform a comprehensive survey analyzing challenges and problems that FL faces when deployed on edge devices \cite{hetersurvey}. They cover topics like communications costs, resource allocation, and privacy and security issues. Of these challenges, resource allocation is most relevant to our study. They argue resource allocation strategy is important to make FL more efficient due to factors like resource constraints, communication bandwidth, and heterogeneity in data and hardware. 

They discuss multiple resource allocation challenges like participant selection (the focus of this study), adaptive aggregation, and joint radio and computation resource management. Finally, they discuss challenges that have not yet been addressed, one of which is called dropped clients. They argue that current selection techniques do not consider the availability and possibility of client dropouts (failures), which needs to be addressed. Our study is aimed to address this exact issue by proposing new factors in the client selection phase of FL.

Another survey covers FL's challenges when deployed on IoT devices with limited resources \cite{imteaj2021survey}. It starts by covering the basics of FL and then discusses resource challenges of FL like communication overhead, heterogeneous hardware, scheduling (client selection), and privacy. Furthermore, they acknowledge the issue of dropped clients due to availability issues.
They say that any of the clients may drop out of the network during the training phase due to different factors like movement, low battery, and low bandwidth. The study suggests that solutions have been proposed to handle straggler clients like asynchronous FL, but like the last survey, they believe the issue needs further investigation. 

As the related surveys suggested, client selection is a very important component in the resource-constrained FL, and many techniques have been proposed other than what was studied in this study.

One of these techniques is called Oort \cite{oort}. Oort works by assigning each client a utility function. The utility function consists of two parts: statistical utility and system utility. The statistical utility is collecting the training loss of samples in each round for each client and is used to make the algorithm fair to ensure accuracy. While the system utility is trying to make the algorithm select faster clients and make the process faster. In each round, this approach selects a fraction of clients based on this utility function (called exploitation) and selects the rest of the required clients randomly (called exploration). This exploration/exploitation ratio changes as the training progress.

According to the authors, this technique does not perform well when the number of clients per round is small, and this could be because the exploration and exploitation fractions here need a certain number of clients to perform as expected. Furthermore, this technique has many design choices that need to be optimized, like exploration/exploitation ratio, utility functions, etc. Since our study uses a small number of clients per round and does not use the same datasets as Oort, we did not include Oort in our study.

RBCS-F is another client selection technique introduced to achieve fast training times while preserving the fairness \cite{efficienyselection}. This technique solves the client selection problem with an approach called Contextual Combinatorial Multi Arm Bandit (C\^2 MAB), in which a subset of arms has to be selected from a set of arms to maximize the total reward. It considers each client to be an arm and assigns a reward to each client based on the time it consumes for training. To ensure fairness is guaranteed, it uses optimization techniques to ensure all clients are selected at least a certain number of times. Though this technique sounds promising, we excluded it from our study since the algorithms are complex, some parameters are unknown, and there is no replication package for this study.

Power-of-Choice is another client selection technique that has been proposed recently \cite{powerofchoice}. Unlike other selectors that we talked about, Power-of-Choice is not trying to decrease the training time by considering the resources. Rather, it aims to achieve higher accuracies in a certain number of rounds. In essence, it samples a set of candidates, then estimates the clients' local losses and selects a subset of clients with the highest loss value. Since the goal of this study is not to decrease the number of training rounds to achieve a certain accuracy, we did not study it in this paper.

Apart from the discussed techniques, other techniques are proposed that are not exactly within the scope of this study. For instance, a group of selectors focuses on asynchronous FL \cite{async1, async2}, and some techniques focus on hierarchical FL \cite{hier1, hier2}. Thus we did not use these techniques in our study.

\section{Conclusion and future work}
\label{conclusion}
In this study, we tackled one of the most important issues in cross-device FL: clients' availability. We analyzed FL under different availability scenarios. We proposed MDA as the first availability-aware client selector that reduces client failures mid-training, causing fewer timeouts in the server and consequently speeding up the learning process. Furthermore, we compared state-of-the-art resource-aware selectors and showed the possibility of having a selector that considers both the availability and resource heterogeneity of clients.

Our results show that low availability can slow down the FL process by up to 10\% compared to a high availability case. Also, it can degrade the accuracy by up to 5\%.
Furthermore, MDA can improve vanilla FL (the Random selector), make the process faster by up to 6.5\%, and decrease the number of timed-out rounds by up to 38\%.
Additionally, results show that TiFL is the fastest state-of-the-art technique and can speed up the learning process by up to 60\%.
Lastly, our approach (TiFL-MDA) that considers both availability and resource heterogeneity can further improve TiFL by up to 16\%, making it the fastest client selector among the studied techniques.

In the future, we plan to extend the TiFL-MDA technique to make it fairer and improve the model accuracy. We intend to use factors like clients' local losses to achieve this goal.

% \begin{acknowledgements}
% Not sure what to put here.
% \end{acknowledgements}

% % BibTeX users please use one of
\bibliographystyle{IEEEtran}
\bibliography{refs}

\end{document}